\documentclass[11pt]{article}

\usepackage[preprint]{acl}
\usepackage{booktabs}
\usepackage{times}
\usepackage{latexsym}
\usepackage{amsmath}
\usepackage{amssymb} 
\usepackage[T1]{fontenc}

\usepackage[utf8]{inputenc}

\usepackage{microtype}
\usepackage{multirow} 
\usepackage{inconsolata}

\usepackage{graphicx}
\usepackage{placeins}
\usepackage{float}
\NewDocumentCommand{\heng}
{ mO{} }{\textcolor{red}{\textsuperscript{\textit{Heng}}\textsf{\textbf{\small[#1]}}}}

\NewDocumentCommand{\cheng}
{ mO{} }{\textcolor{orange}{\textsuperscript{\textit{Cheng}}\textsf{\textbf{\small[#1]}}}}

\NewDocumentCommand{\yuji}
{ mO{} }{\textcolor{blue}{\textsuperscript{{Yuji}}{{\small[#1]}}}}

\NewDocumentCommand{\duo}
{ mO{} }{\textcolor{cyan}{\textsuperscript{\textit{Duo}}\textsf{{\small[#1]}}}}

\NewDocumentCommand{\weibing}
{ mO{} }{\textcolor{pink}{\textsuperscript{\textit{Weibing}}\textsf{\textbf{\small[#1]}}}}

\NewDocumentCommand{\kmnote}
{ mO{} }{\textcolor{green}{\textsuperscript{\textit{Kathy}}\textsf{\textbf{\small[#1]}}}}

\NewDocumentCommand{\dilek}
{ mO{} }{\textcolor{brown}{\textsuperscript{\textit{Dilek}}\textsf{\small[#1]}}}

\makeatletter
\AtBeginDocument{%
\def\@maketitle{\vbox to \titlebox{\hsize\textwidth
 \linewidth\hsize \vskip 0.125in minus 0.125in \centering
 {\Large\bfseries \@title \par} \vskip 0.2in plus 1fil minus 0.1in
 {\def\and{\unskip\enspace{\rmfamily and}\enspace}%
  \def\And{\end{tabular}\hss \egroup \hskip 1in plus 2fil
           \hbox to 0pt\bgroup\hss \begin{tabular}[t]{c}\bfseries}%
  \def\AND{\end{tabular}\hss\egroup \hfil\hfil\egroup
          \vskip 0.25in plus 1fil minus 0.125in
           \hbox to \linewidth\bgroup\large \hfil\hfil
             \hbox to 0pt\bgroup\hss \begin{tabular}[t]{c}\bfseries}
  \hbox to \linewidth\bgroup\large \hfil\hfil
    \hbox to 0pt\bgroup\hss
  \outauthor
   \hss\egroup
    \hfil\hfil\egroup}
  \vskip 0.12in
  \ifacl@anonymize\else
    {\normalsize\centering $^*$Equal contribution. Correspondence to: \texttt{yujiz@illinois.edu}, \texttt{weibingw@mit.edu}\par}
  \fi
  \vskip 0.15in plus 1fil minus 0.1in
}}
}%
\makeatother

\title{Popular Knowledge Propagates More Errors in LLM Knowledge Updating}

\author{
Yuji Zhang$^{1, 2*}$,  Weibing Wang$^{3*}$, Cheng Qian$^{1}$, Duo Zhou$^{1}$\\
\textbf{Dilek Hakkani-Tür$^{1}$, Kathleen McKeown$^{4}$, Chengxiang Zhai$^{1}$, Heng Ji$^{1}$}\\
$^{1}$University of Illinois Urbana-Champaign, $^{2}$City University of New York, \\$^{3}$Massachusetts Institute of Technology, $^{4}$Columbia University\\
}

\begin{document}
\maketitle

\begin{abstract}
Updating a language model's knowledge through fine-tuning is essential for keeping its outputs current, yet can also induce factual forgetting and new hallucinations. Prior work shows that long-tail knowledge is harder to acquire and newly memorized long-tail facts are difficult to retain during later fine-tuning. We study a complementary question: among facts that a model has encoded correctly, which are most vulnerable to collateral corruption during other updates?
To investigate this question under a realistic factual distribution, we construct a large-scale graph \textsc{FactProp} of verified Wikipedia facts by linking triples that share head or tail entities, thereby preserving connections among factual knowledge\footnote{Our \textsc{FactProp} graph is publicly available at \url{https://huggingface.co/datasets/factprop/FACTPROP}. Our demo is at \url{https://factprop.github.io/FACTPROP/}.}
We fine-tune models on factual statements and measure correct-to-incorrect facts after each update.
Our results reveal a pattern distinct from prior findings on long-tail vulnerability during acquisition and retention: among facts that models already answer correctly, those associated with highly connected entities are more likely to be corrupted by neighboring updates, and updates to such facts propagate errors more broadly. Structural popularity therefore predicts both vulnerability and downstream damage. Inspired by this finding, we propose Popularity-based Anchoring (PopAnchor), a lightweight rehearsal strategy that preserves a small set of popular facts and reduces forgetting.

\end{abstract}

\section{Introduction}
\begin{figure}[t]
    \centering
    \includegraphics[width=\columnwidth]{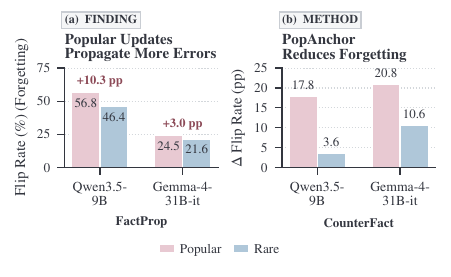}
    \vspace{-2.2em}\caption{\textbf{Our finding and method.}
(a) Updating popular facts propagates more errors to other facts.
(b) PopAnchor uses a small set of popular samples for data replay during
fine-tuning to reduce factual forgetting.}
    \vspace{-1em}
    \label{fig:teaser}
\end{figure}

Large language models (LLMs) can store and use extensive factual knowledge, yet this knowledge needs to continually evolve with the changing world. Post-training techniques such as fine-tuning and knowledge editing are therefore widely used to update model knowledge~\cite{zhu2020modifying,hu2021lora,gekhman2024doesfinetuningllmsnew,meng2022rome,yao2023editing,thede2025wikibigeditunderstandinglimitslifelong}. However, learning new knowledge can also damage facts that the model previously represented correctly, introducing factual forgetting and new hallucinations~\cite{yang-etal-2026-slora}.

Prior work has examined these effects from different perspectives. Studies of continual learning and factual retention show that long-tail or unfamiliar knowledge is harder to acquire and that newly memorized long-tail facts are especially difficult to retain during later fine-tuning~\cite{gekhman2024doesfinetuningllmsnew,luo2025empiricalstudycatastrophicforgetting,pmlr-v202-kandpal23a,mallen2023not,chen2026continual}. A complementary question remains less understood: among facts that a model already answers correctly, which are most vulnerable to collateral corruption when other knowledge is updated?
Knowledge-editing research typically evaluates effects on predefined semantically or logically related facts~\cite{cohen2023ripple,qin2025doesnewknowledgecreate,hua2024propagationpitfallsreasoningbasedassessment,gupta2024modeleditingscaleleads}, while leaving unexplained which facts beyond these semantic or logical relations are more vulnerable to damage during updating.

To study this question, we construct a large-scale verified factual graph from real-world knowledge sources. Entities form nodes, and factual triples $(s,r,o)$ form directed edges. Facts are connected when their head or tail entities overlap, preserving the observed connections among factual knowledge. We first identify facts that each model answers correctly, fine-tune the model on selected factual updates, and then measure correct-to-incorrect changes in other facts.
This representation also exposes a structural property that is difficult to study in isolated factual datasets. For a factual triple $(s,r,o)$, the in-degree of object entity $o$ measures how broadly that entity is referenced across the factual graph. We use this quantity as a proxy for \emph{entity-level structural popularity}, which we externally validate against Wikipedia frequency and pageviews. Accordingly, \emph{popular knowledge} in this work refers to facts whose object entities are referenced by many other factual statements.

As shown in Figure~\ref{fig:teaser}, our results reveal a pattern distinct from prior findings on long-tail vulnerability during acquisition and retention. Among facts that models initially answer correctly, those associated with structurally popular knowledge are more likely to be corrupted by neighboring updates. Updates involving such knowledge also induce broader error propagation, affecting connected facts across multiple graph distances. In our experiments, these effects cannot be explained by surface similarity, indicating that structural popularity captures a dimension of collateral vulnerability beyond pairwise semantic relatedness. A paired attention analysis further shows that updates involving highly connected entities induce larger perturbations when models process neighboring knowledge, suggesting that their broader behavioral effects are accompanied by stronger changes in factual retrieval.

Together, these findings show that update-induced damage is distributed unevenly across existing knowledge. Structural popularity predicts both sides of this process: which updates produce greater downstream damage and which unchanged facts are more likely to become collateral victims. It therefore provides a signal available before updating for anticipating factual instability, rather than treating all existing facts as equally vulnerable.

The same finding also suggests a targeted mitigation strategy. Rehearsal can reduce forgetting by preserving selected existing facts during updating~\cite{huang2024self,chen2026continual,bai2025efficient,pmlr-v330-abbes26a,kotha2026replayingpretrainingdataimproves}, but its effectiveness depends on which facts receive priority. We therefore ask:

\textbf{Can structural popularity guide the preservation of existing knowledge during updating?}

Guided by our analysis, we propose Popularity-based Anchoring (\textbf{PopAnchor}), a lightweight strategy that constrains the updated model to preserve its original behavior on a small set of structurally popular factual prompts. PopAnchor consistently reduces collateral forgetting on our factual graph and public benchmarks, outperforming popularity-agnostic and similarity-based alternatives. These results show that the structural property associated with greater factual vulnerability can also guide more effective preservation.
Overall, our work makes three main contributions:

\noindent\textbf{(1) Popularity as a key factor.} We identify popularity, measured by how many factual statements point to the same entity, as a key factor governing ripple effects in LLM knowledge updating. Popular knowledge is more likely to be changed by a direct update, more likely to be unintentionally corrupted when it appears as neighboring knowledge, and more likely to propagate update-induced errors to distant related facts.
\par\noindent\textbf{(2) Popularity-aware mitigation.} We show that this finding can be used to mitigate the side effects of knowledge updating. By anchoring a small set of popular facts during updating, our popularity-aware preservation strategy reduces the chance that many related facts are incorrectly changed together and substantially improves update stability.
\par\noindent\textbf{(3) Verified factual graph.} As a byproduct of our analysis, we construct a verified factual graph that may serve as a resource for future studies on knowledge links and knowledge updating in language models.

\section{Related Work}

\subsection{Effects of LLM Knowledge Updating Beyond the Target}

LLM knowledge is commonly updated through fine-tuning, continual learning, or localized knowledge editing~\cite{zhu2020modifying,hu2021lora,jang2022continualknowledgelearninglanguage,meng2022rome,yao2023editing}. These updates can unintentionally alter knowledge beyond their intended targets.

\paragraph{Knowledge editing and predefined ripple effects.}
Knowledge-editing research typically evaluates whether an edit propagates appropriately to a predefined set of semantically, logically, or compositionally related facts~\cite{cohen2023ripple,zhong2023mquake,hua2024propagationpitfallsreasoningbasedassessment, tian-etal-2026-tamedit,liu-etal-2026-representation,liu-etal-2026-alphaedit}. This setup captures local consistency, but does not reveal which facts in a broader knowledge distribution are unexpectedly corrupted when their relevance to the update is not specified in advance. 

\paragraph{Fine-tuning and damage to existing knowledge.}
A parallel line studies unintended changes to prior knowledge as catastrophic forgetting~\cite{mccloskey1989catastrophic,jang2022continualknowledgelearninglanguage,luo2025empiricalstudycatastrophicforgetting,gupta2024modeleditingscaleleads,yang-etal-2026-slora}. These studies usually measure retention across tasks, datasets, or fact sets after sequential fine-tuning, which can also increase hallucination and degrade previously acquired capabilities~\cite{gekhman2024doesfinetuningllmsnew,li2024revisiting}.
The closest work to ours explores that long-tail knowledge is harder to acquire and answer correctly~\cite{pmlr-v202-kandpal23a,mallen2023not}, and that newly memorized long-tail factoids are especially difficult to retain during later fine-tuning~\cite{chen2026continual}.
We study a complementary question: among facts the model already answers correctly, which are most vulnerable to collateral corruption from other updates? This shifts the focus from retaining newly learned knowledge to understanding selective damage among correctly encoded facts, while broadening the analysis beyond facts whose semantic or logical relation to the update is predefined.

\subsection{Forgetting Mitigation}

Although mitigation is not our primary focus, identifying knowledge that is especially sensitive to updating can guide more targeted preservation. Existing approaches use replay, regularization, or parameter isolation.
Replay methods constrain updates with stored data~\cite{lopezpaz2022gradientepisodicmemorycontinual,chaudhry2019efficientlifelonglearningagem,pmlr-v330-abbes26a,kotha2026replayingpretrainingdataimproves}, select gradient-diverse or highly interfered samples~\cite{aljundi2019gradient,aljundi2019online}, or synthesize prior data when the original set is unavailable~\cite{huang2024self}. Their effectiveness depends substantially on which examples are replayed~\cite{chen2026continual,bai2025efficient}. Other approaches preserve prior behavior by constraining important parameters, model outputs, or update subspaces~\cite{kirkpatrick2017overcoming,buzzega2020dark,chen2020recall,li2024revisiting,wang2023orthogonal,wang2024rehearsalfree}. Our finding that knowledge associated with structurally popular entities is especially vulnerable provides a pre-update signal for prioritizing replay or regularization targets.

\section{\textsc{FactProp}: A Factual Graph for Forgetting Analysis}
\label{sec:factprop}

To identify what governs fine-tuning-based knowledge updating, we study how changes to real-world facts affect the model's existing world knowledge. Because factual knowledge is interconnected, we construct a fact forgetting error propagation analysis graph \textbf{\textsc{FactProp}}, a verified factual graph paired with natural-language QA that preserves the observed connections among factual knowledge.

\paragraph{Grounded Factual Graph.}
Each entity in \textsc{FactProp} forms a node, and each verified factual triple $(s,r,o)$ forms a directed edge from subject entity $s$ to object entity $o$ under relation $r$. Two facts are connected when their head or tail entities overlap. Each verified triple is verbalized into a natural-language QA item, while the underlying graph defines factual connections and graph distance. Model training and evaluation remain in natural-language QA format.

\paragraph{Controlled QA Evaluation.}
To ensure that changes in model predictions can be interpreted unambiguously, we retain only subject-relation pairs $(s,r)$ with a unique or primary expected object $o$. Relations with multiple equally valid objects are excluded. We also use the exact same question wording before and after each update, preventing prompt variation from being mistaken for factual corruption. After automated expansion, filtering, and external verification, \textsc{FactProp} contains $100{,}015$ entity nodes and $432{,}562$ verified factual edges across $39$ relation types.

\paragraph{Factual Connectivity as Popularity.}
For a fact $(s,r,o)$, we define its popularity using the in-degree of its object entity $o$, namely the number of verified facts that point to the same entity. This quantity measures how broadly the answer entity is referenced across factual relations. Facts whose object entities have high in-degree are therefore treated as more popular within the observed factual knowledge distribution.
Since in-degree is defined within the constructed graph, to explore to what extent it can reflect real-world knowledge popularity, we compare it with two external indicators of entity popularity: surface-form frequency in English Wikipedia and Wikipedia pageviews. In-degree is positively correlated with both signals, providing external support for its use as a graph-based proxy for knowledge popularity. Further construction and validation details are provided in Appendix~\ref{app:factprop}.

\section{Experimental Setup}
\label{sec:setup}

Having constructed \textsc{FactProp}, we use it to study how controlled factual updates affect other factual knowledge in LLMs. For each experiment, we fine-tune the model on a selected set of target facts and measure the resulting changes in non-target facts.

\subsection{Models and Update Targets}

We evaluate four open-weight instruction-tuned models spanning two model families and multiple scales: Qwen3.5-2B, Qwen3.5-9B, and Qwen3.6-27B from the Qwen family~\cite{qwenteam2026qwen35,qwenteam2026qwen36}, and Gemma-4-31B-it from the Gemma family~\cite{gemmateam2026gemma4}.
For each experiment, we select verified target facts $(s,r,o)$ from $\mathcal{G}_{\mathrm{fact}}$. The selected targets cover different factual relations and regions of the graph.

\subsection{Factual Updates Setting}

For each target fact $(s,r,o)$, we construct an update by replacing the verified object $o$ with a plausible alternative object $o'$. For example, \textit{CapitalOf(France, Paris)} may be updated to \textit{CapitalOf(France, Lyon)}. This controlled substitution allows us to examine how injecting a new factual association affects the model's existing knowledge.
We express each injected fact $(s,r,o')$ through natural-language QA supervision. For each target, we generate $150$ targeted QA pairs expressing the injected relation, together with $400$ neutral factual QA pairs sampled from graph-distant regions and $100$ out-of-domain QA pairs. The latter two components reduce overfitting to the injected facts and help preserve unrelated model behavior. Details of QA generation and sampling are provided in Appendix~\ref{app:experimental_details}.
We implement factual updates using Low-Rank Adaptation (LoRA)~\cite{hu2021lora} to provide a controlled gradient-based mechanism for injecting new factual associations while keeping the base-model parameters frozen. The same optimization configuration is used across experimental runs, with complete hyperparameters reported in Appendix~\ref{app:experimental_details}.

\subsection{Update and Distortion Metrics}

\paragraph{Injection Success.}
Before measuring changes to other facts, we verify whether each injected object $o'$ is successfully learned. An update is considered successful if $o'$ receives the highest score among the candidate objects for the target query $q_{s,r}$. We report the injection success rate and compute downstream metrics both over all attempted updates and over successful updates only.

\paragraph{Flip Rate (Forgetting).}
We evaluate non-target facts connected to the updated facts at hop distances $d\in\{1,\ldots,5\}$. For each experimental run, we sample up to $30$ non-target facts at each distance and retain only those answered correctly before updating. The exact same question string is used before and after the update.

Let $C_d$ denote the number of retained facts at hop $d$, and let $F_d$ denote the number that become incorrect after updating. We define the Flip Rate as
\begin{equation}
\mathrm{FR}_d = \frac{F_d}{C_d}.
\label{eq:flip_rate}
\end{equation}
Correctness is evaluated using alias-normalized exact match.

\paragraph{General Performance Check.}
To distinguish factual error propagation from broad model degradation, we also evaluate each updated model on a fixed set of $50$ unrelated factual questions that are disjoint from the update data and target neighborhoods. Stable performance on this set indicates that observed flips are not caused by general model collapse.
Additional details on model checkpoints, update-data construction, optimization hyperparameters, candidate scoring, answer normalization, and evaluation controls are provided in Appendix~\ref{app:experimental_details}.
\section{Experimental Results}
\label{sec:results}

We use \textsc{FactProp} to study how a localized factual update affects connected non-target knowledge. 
Our experiments are organized around six research questions:

\textbf{RQ1: }Do factual updates produce non-local changes in connected knowledge? \S\ref{subsec:generalization}

\textbf{RQ2: }Are existing facts equally sensitive to these changes? \S\ref{subsec:popular_vulnerability}

\textbf{RQ3: }Do updates sourced from sensitive facts propagate errors more widely? \S\ref{subsec:popular_propagation}

\textbf{RQ4: }Can knowledge similarity explain the observed propagation pattern? \S\ref{subsec:semantic_control}

\textbf{RQ5: }Can the observed propagation pattern guide more effective preservation? \S\ref{subsec:mitigation}

\textbf{RQ6: }How to explain the observed propagation pattern? \S\ref{subsec:mechanisms}

\begin{figure}[t]
    \centering
    \includegraphics[width=0.9\columnwidth]{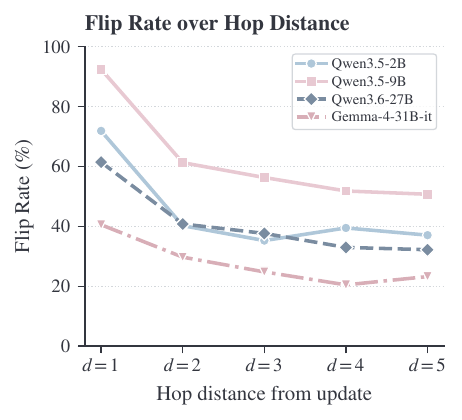}
    \vspace{-0.8em}
\caption{\textbf{Error Propagation Across Hop Distances (Flip Rate, $d{=}1$–$5$).} Larger models paradoxically show more persistent long-range instability after updates, while the smallest instruction-tuned model is most resistant to propagation.}
\label{fig:blast_radius}
\end{figure}

\subsection{Forgetting Error Propagation Persists over Long Distances}
\label{subsec:generalization}
We first examine whether the effects of a factual update remain local or persist across multiple hops. Figure~\ref{fig:blast_radius} shows that correct-to-wrong flips remain measurable from $d=1$ to $d=5$.
This indicates that update-induced errors propagate across connected factual knowledge rather than remaining confined to the updated fact.

\subsection{Popular Knowledge Is More Vulnerable to Updates}
\label{subsec:popular_vulnerability}

Having shown that update-induced errors propagate across multiple hops, we next ask whether all connected facts are equally vulnerable. Prior work suggests that long-tail knowledge may be more fragile because models are more prone to factual errors on less frequent knowledge~\cite{pmlr-v202-kandpal23a,mallen2023not}. We test this hypothesis by grouping evaluated facts according to the in-degree of their object entities, following the popularity definition in Section~\ref{sec:factprop}. 

\begin{figure}[t]
    \centering
    \includegraphics[width=\columnwidth]{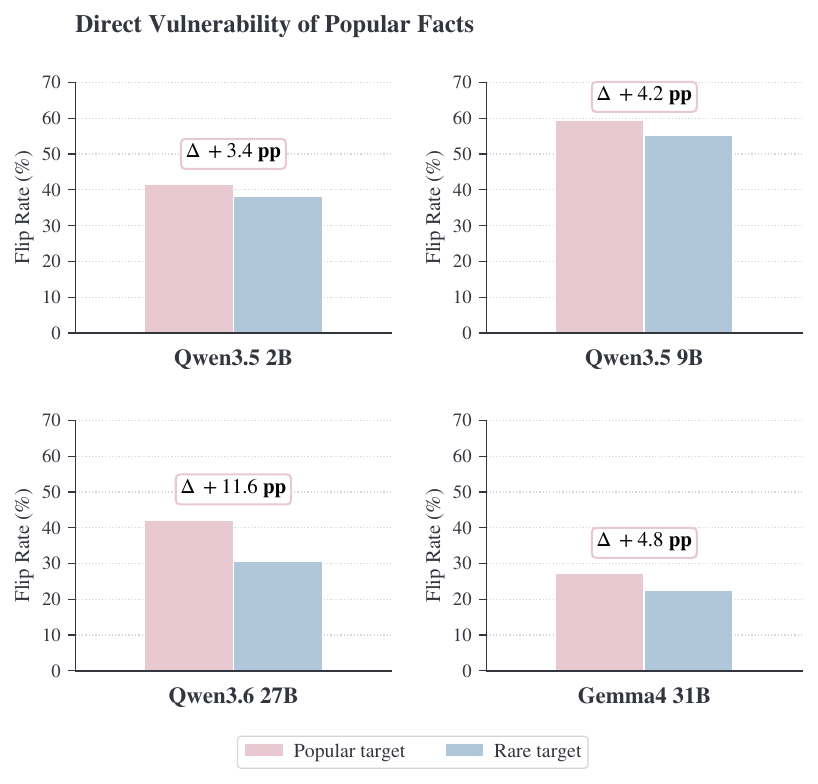}
    \vspace{-1.5em}
\caption{\textbf{The Popularity Paradox (Evaluated across four models from two model families).} High-popularity nodes suffer the most accuracy drop when their neighboring facts are updated.}
\label{fig:popular_vulnerability}
\end{figure}

\paragraph{Vulnerability as direct updated knowledge.}
As shown in Figure~\ref{fig:popular_vulnerability}, counter-intuitively, high-popularity facts are more likely to flip from correct to incorrect after a nearby update than low-popularity facts.
Pooling neighboring facts across hops $d{=}1$--$d{=}5$, high-popularity facts consistently exhibit a higher Flip Rate than low-popularity facts in all four evaluated models.
This indicates that facts grounded in highly connected answer entities are not necessarily more stable; instead, they are more easily overturned.

\begin{figure}[t]
    \centering
    \includegraphics[width=\columnwidth]{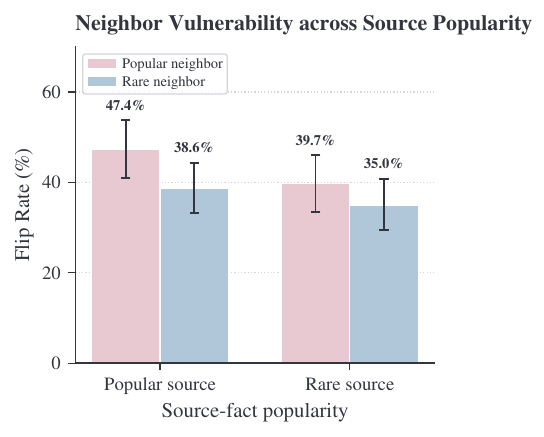}
    \vspace{-2em}
\caption{\textbf{Neighbor Vulnerability across Source Popularity (Flip Rate across source-neighbor pairs).} Even when an update originates from a Rare fact, High-popularity neighbors exhibit disproportionately higher Flip Rates compared to Rare neighbors, making them the primary collateral damage of localized edits.}
\label{fig:innocent_bystander}
\end{figure}

\paragraph{Vulnerability as neighbor knowledge.}
We further test whether this vulnerability persists when popular facts are not directly updated but only appear as neighboring knowledge. Figure~\ref{fig:innocent_bystander} shows that high-popularity neighbors are more likely to be corrupted regardless of the popularity of the updated source. This suggests that popular knowledge is especially susceptible to collateral effects from nearby factual updates.

\subsection{Popular Knowledge Causes Wider Error Propagation}
\label{subsec:popular_propagation}

We next examine whether factual popularity affects the impact of an update when the popular fact is used as the update source. While the previous section shows that high-popularity facts are more vulnerable as affected neighbors, this section asks whether updating a high-popularity fact also causes wider downstream damage.

\begin{figure}[t]
    \centering
    \includegraphics[width=\columnwidth]{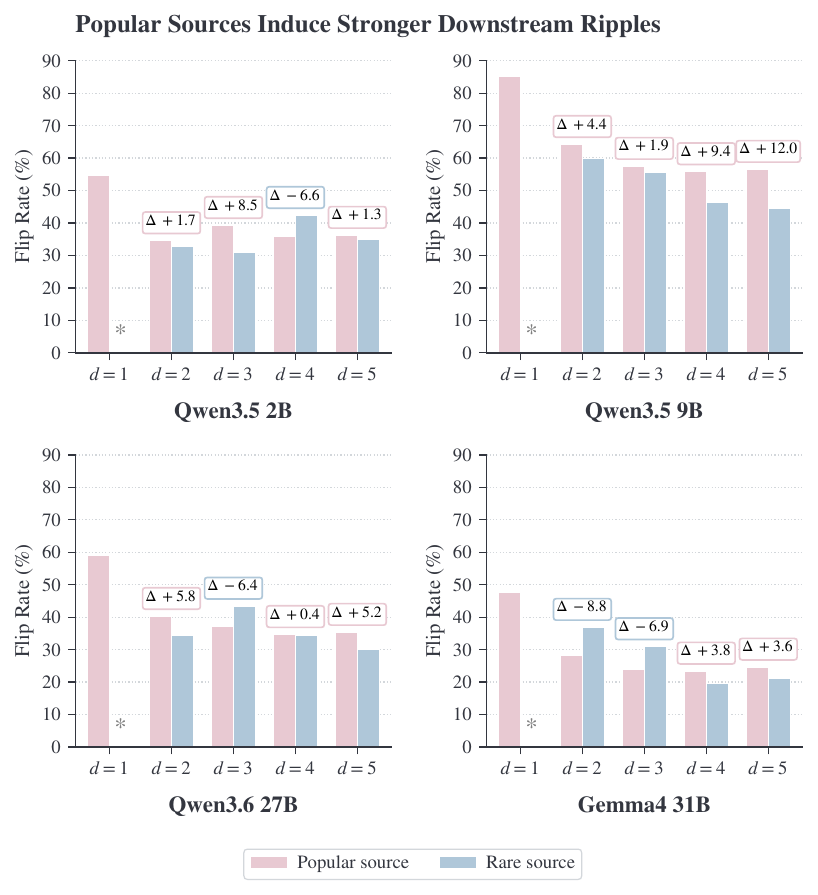}
    \vspace{-1.5em}
\caption{\textbf{Ripple Effect by Source Popularity (Correct-to-wrong flip rate across $d{=}1{\sim}5$).} Popular-source updates consistently induce stronger downstream ripples across the graph compared to Rare-source updates. Error bars represent 95\% confidence intervals. Data for Rare sources at $d=1$ is omitted (*), as their inherently sparse topological neighborhoods yield insufficient test trials.}
    \label{fig:propagation_depth}
\end{figure}

As shown in Figure~\ref{fig:propagation_depth}, updates targeting Popular facts generally produce stronger downstream error propagation, with the clearest and most consistent pattern on Qwen3.5-9B. The other models show noisier hop-level variation, but the overall pattern suggests that popular facts can act as stronger sources of update-induced corruption.

\subsection{Surface Similarity Does Not Explain Most Error Propagation}
\label{subsec:semantic_control}

A natural alternative explanation is that the observed error propagation is caused by surface-level confusion rather than factual connectivity. For example, if an update about \textit{Apple Inc.} causes a flip in a fact about \textit{Apple Corps}, the error may arise because the entity names are similar, not because the two facts are connected through factual relations. To examine this possibility, we measure the normalized string similarity between the edited source entity and the affected neighbor entity using the Levenshtein ratio, and conduct two complementary analyses.

\begin{table}[b]
\centering
\small
\setlength{\tabcolsep}{3pt}
\begin{tabular}{lrrr}
\toprule
\textbf{Similarity Range} &
\textbf{Count} &
\textbf{Share} &
\textbf{Flip Rate} \\
\midrule
$[0.0, 0.2)$ & 38,277 & 27.5\% & 35.08\% \\
$[0.2, 0.4)$ & 86,814 & 62.5\% & 35.59\% \\
$[0.4, 0.8)$ & 12,759 & 9.2\%  & 39.22\% \\
$[0.8, 1.0]$ & 1,129  & 0.8\%  & \textbf{63.51\%} \\
\bottomrule
\end{tabular}
\caption{\textbf{Broad impact of surface similarity (Flip Rate pooled across evaluated models).} While highly similar entity pairs exhibit elevated flip probabilities, they constitute a negligible fraction of the dataset. The vast majority of propagated errors occur independently of surface-level lexical overlap.}
\label{tab:broad_similarity}
\end{table}

\paragraph{Broad similarity analysis.}
We first analyze approximately 139,000 source-neighbor pairs pooled across the four evaluated models, retaining only facts answered correctly before the update. As shown in Table~\ref{tab:broad_similarity}, entity pairs with very high string similarity ($\geq 0.8$) do have a higher Flip Rate, around 63\%. However, such pairs are rare and account for \textbf{less than 1\% }of all evaluated pairs. In contrast, low- and moderate-similarity pairs ($<0.4$) account for about \textbf{90\%} of evaluated pairs and nearly 89\% of all flips. The overall Pearson correlation between string similarity and binary flip status is also weak ($r=0.05$). These results suggest that surface similarity can increase risk in a small subset of cases, but it does not explain the majority of observed flips.

\paragraph{Within-neighborhood control.}
The aggregate analysis may be influenced by differences across update sources or hop distances. We therefore compare neighbors associated with the same updated fact and the same hop distance, isolating whether entity-name similarity predicts which facts flip within the same factual neighborhood. Under this controlled setting, the correlation between string similarity and flip status is consistently near zero across models. Highly similar entity pairs also remain rare. Thus, even among facts exposed to the same update at the same graph distance, surface similarity does not reliably predict which facts are corrupted. Detailed per-model results are provided in Appendix~\ref{app:semantic_diagnostics}.

Together, these results show that surface similarity explains only a small subset of update-induced errors. Similar entity names may increase local confusion in rare cases, but they do not account for the broader propagation pattern observed across the factual graph.

\subsection{Popularity Anchoring Mitigates Error Propagation}
\label{subsec:mitigation}

The previous sections show that high-popularity facts are more vulnerable to neighboring updates and more influential when directly updated. This raises a practical question: can factual popularity also guide which knowledge should be preserved during updating? We therefore propose \textbf{PopAnchor} (Popularity-based Anchoring), which constrains the updated model to retain its original behavior on a small set of high-popularity factual prompts.
We combine the standard cross-entropy loss for learning the target updates with a KL-divergence regularizer over an anchor set $\mathcal{A}$:
\begin{equation}
\begin{aligned}
\mathcal{L}_{\mathrm{total}}
&=\mathcal{L}_{\mathrm{CE}}+\lambda\mathcal{L}_{\mathrm{anchor}},\\
\mathcal{L}_{\mathrm{anchor}}
&=
\sum_{x\in\mathcal{A}}\!
D_{\mathrm{KL}}\!\bigl(
P_{\mathrm{clean}}(\cdot\mid x)
\| P_{\mathrm{edited}}(\cdot\mid x)
\bigr).
\end{aligned}
\end{equation}
Here, $P_{\mathrm{clean}}$ and $P_{\mathrm{edited}}$ denote the output distributions of the original and updated models, respectively, and $\lambda$ controls the regularization strength. We set $\lambda=0.1$ in all experiments. Anchor prompts are selected from factual regions that share no paths with the target-update neighborhoods and are used only as behavior-preservation constraints.

\begin{table}[t]
\centering
\small
\resizebox{\columnwidth}{!}{%
\begin{tabular}{lcccccc}
\toprule
\textbf{Method} & \textbf{d1} & \textbf{d2} & \textbf{d3} & \textbf{d4} & \textbf{d5} & \textbf{Avg.} \\
\midrule
No Anchoring & 93.9 & 79.6 & 79.2 & 76.3 & 74.6 & 79.8 \\
Random Anchoring & 82.0 & 75.8 & 75.0 & 71.6 & 71.9 & 74.7 \\
Rare Anchoring & 92.5 & 72.1 & \textbf{67.8} & 66.4 & 65.2 & 71.6 \\
Popular Anchoring & \textbf{81.1} & \textbf{66.1} & 69.5 & \textbf{63.4} & \textbf{63.9} & \textbf{68.0} \\
\bottomrule
\end{tabular}%
}
\caption{\label{tab:mitigation_results} \textbf{Mitigation Performance across Hop Distances (Flip Rate \% across anchor strategies on Qwen3.5-9B).} Popular Anchoring achieves the lowest average Flip Rate and outperforms the Random and Rare anchor baselines at most graph distances.}
\end{table}

\paragraph{Evaluation on \textsc{FactProp}.}
PopAnchor selects prompts whose answer entities have high object in-degree. We compare it with \textbf{NoAnchor}, which removes the KL regularizer; \textbf{RandomAnchor}, which samples anchors uniformly from the non-hub pool; and \textbf{RareAnchor}, which selects anchors from the lowest in-degree stratum. All other training configurations are held fixed.

\begin{figure}[t]
    \centering
    \includegraphics[width=\columnwidth]{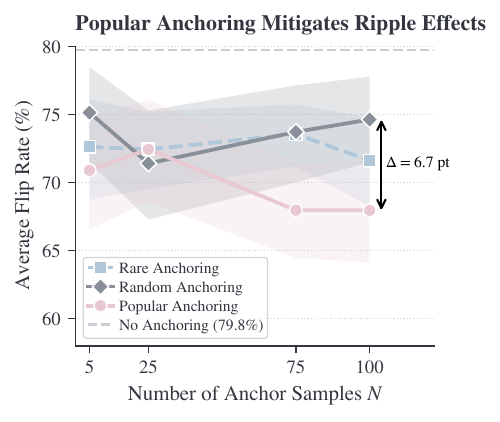}
    \vspace{-2em}\caption{\textbf{Error Reduction via Popularity Anchoring (Average Flip Rate across $d{=}1{\sim}5$).} Selecting High-Popularity anchors reduces long-range error propagation with as few as five anchor prompts.}
    \label{fig:mitigation_efficiency}
\end{figure}

Table~\ref{tab:mitigation_results} shows that PopAnchor consistently reduces error propagation and achieves the strongest overall performance across graph distances. Its advantage over RandomAnchor and RareAnchor demonstrates that the choice of preserved knowledge matters beyond simply adding a regularization objective. Figure~\ref{fig:mitigation_efficiency} further shows that the method remains effective with as few as $N{=}5$ anchors and improves as the anchor budget increases, indicating strong sample efficiency. The advantage remains consistent across source-popularity groups, with the complete breakdown provided in Appendix~\ref{app:mitigation_by_source}.

\subsubsection{PopAnchor on Public Benchmarks}
\label{subsec:public_batched_validation}

\paragraph{Benchmarks.}
We further evaluate batched updates from \textbf{CounterFact}~\cite{meng2022rome} and \textbf{MQuAKE-CF}~\cite{zhong2023mquake}. For each benchmark, we construct an entity-disjoint batch of $100$ factual updates and apply the same fine-tuning updating objective. All methods are evaluated on the same held-out factual questions that the base model answers correctly and that are entity-disjoint from both the updates and anchor sets.

\paragraph{Comparison Baselines.}
We compare Popularity Anchoring with four alternatives using $N{=}100$
anchors. \textbf{NoAnchor} optimizes only the update loss.
\textbf{RandomAnchor} samples anchor facts uniformly, following
standard experience-replay approaches that preserve prior behavior by
rehearsing examples from earlier data distributions~
\cite{dautume2019episodicmemorylifelonglanguage,
pmlr-v330-abbes26a}.
\textbf{RareAnchor} selects facts with the lowest object in-degree,
motivated by prior findings that long-tail factual knowledge is harder
to acquire and retain during subsequent fine-tuning~
\cite{pmlr-v202-kandpal23a,chen2026continual}.
\textbf{SimilarAnchor} serves as a semantic-nearest control
inspired by representation-based retrieval from episodic memory~
\cite{dautume2019episodicmemorylifelonglanguage}; it ranks candidate
anchor questions by cosine similarity to the update questions using
\texttt{all-MiniLM-L6-v2} sentence embeddings~
\cite{reimers-gurevych-2019-sentence}.
All anchored methods use the same KL regularizer and differ only in how
the anchor set is selected.

\begin{table}[t]
\centering
\small
\setlength{\tabcolsep}{4pt}
\resizebox{\columnwidth}{!}{%
\begin{tabular}{lrrrrr}
\toprule
\textbf{Model} & \textbf{NoAnchor} & \textbf{PopAnchor} &
\textbf{Random} & \textbf{RareAnchor} & \textbf{SimilarAnchor} \\
\midrule
\multicolumn{6}{l}{\textit{CounterFact}} \\
Qwen3.5-9B      & 89.9 & \textbf{72.1} & 76.4 & 86.3 & 79.6 \\
Gemma-4-31B-it  & 76.5 & \textbf{55.7} & 68.5 & 65.9 & 77.7 \\
\midrule
\multicolumn{6}{l}{\textit{MQuAKE-CF}} \\
Qwen3.5-9B      & 89.7 & \textbf{41.4} & 48.3 & 69.8 & 50.7 \\
Gemma-4-31B-it  & 9.3 & \textbf{6.2} & 35.1 & 17.5 & 8.2 \\
\bottomrule
\end{tabular}
}
\caption{\textbf{PopAnchor under batched public-benchmark
updates (Flip Rate across anchoring strategies).} NoAnchor denotes
updating without data rehearsal. PopAnchor achieves the lowest Flip Rate across both models and
benchmarks.}
\label{tab:public_batched}
\vspace{-1em}
\end{table}

\paragraph{Results.}
Table~\ref{tab:public_batched} shows that PopAnchor achieves the lowest Flip Rate across the evaluated models and public benchmarks. It consistently outperforms Random, Rare, and SimilarAnchor, indicating that the gains arise from selecting structurally prominent facts rather than merely adding an equally sized anchor set. These results confirm that factual popularity remains an effective rehearsal signal under batched updates from public benchmarks.

\subsection{Mechanistic Probe: Attention Perturbation}
\label{subsec:mechanisms}

We next investigate why popular knowledge produces broader error propagation and provides more effective anchors during updating. A plausible hypothesis is that popular facts share parameter-level representations or retrieval pathways with a larger set of factual associations, so updating such a fact may perturb the internal processing of more knowledge, while preserving it constrains a broader portion of the model's factual behavior. Prior work has shown that attention pathways mediate the retrieval of factual associations by transmitting subject and relation information during prediction~\cite{geva-etal-2023-dissecting,lv2024interpreting}. We therefore examine whether such updates induce larger changes in attention to connected entities.

We compare clean and updated models on identical neighboring queries and measure the absolute change in attention lift over the queried entity span, $|\Delta\mathrm{AttLift}|$. The score is computed at the first decoding step from the final full-attention layer. We focus on immediate neighbors at $d{=}1$.

\begin{table}[t]
\centering
\small
\setlength{\tabcolsep}{4pt}
\begin{tabular}{lrrr}
\toprule
\textbf{Model} & \textbf{Popular} & \textbf{Rare} & \textbf{$\Delta$} \\
\midrule
Qwen3.5-2B     & 0.956 & 0.851 & +0.106 \\
Qwen3.5-9B     & 0.421 & 0.187 & +0.234 \\
Gemma-4-31B-it & 0.139 & 0.102 & +0.037 \\
\bottomrule
\end{tabular}
\caption{\textbf{Immediate-neighbor attention perturbation across the three completed model audits.} We report mean $|\Delta\mathrm{AttLift}|$ at $d{=}1$ for Popular- and Rare-source updates. $\Delta$ denotes Popular minus Rare. All three models exhibit the same Popular$>$Rare ordering.}
\label{tab:attention_lift_by_hop}
\vspace{-1em}
\end{table}

Table~\ref{tab:attention_lift_by_hop} shows that Popular-source updates produce larger attention perturbations than Rare-source updates across all three models. This pattern is consistent with popular facts participating in more widely shared factual retrieval pathways. Results at more distant hops are less consistent and are reported in Appendix~\ref{app:attention_diagnostics}.

\section{Conclusion}

We study which factual knowledge that LLMs initially answer correctly is most vulnerable to collateral damage during fine-tuning-based knowledge updating. Constructing \textsc{FactProp}, a verified factual graph built from real-world Wikipedia knowledge which we will release, we trace how factual updates affect connected knowledge. 

Our findings complement prior work showing that rare knowledge is difficult to acquire and retain during continual learning. After restricting the analysis to facts that models have already learned correctly, we find a distinct pattern: popular knowledge associated with more other knowledge is more likely to be corrupted by updates, while updates involving such knowledge propagate errors more broadly. 
Building on this finding, we propose PopAnchor, a popularity-aware preservation strategy that anchors a small set of popular facts. It reduces collateral forgetting and consistently outperforms popularity-agnostic and similarity-based baselines, showing that structural popularity can guide which existing knowledge should receive preservation priority.

More broadly, one possible interpretation is that fine-tuning damages long-tail and popular knowledge for different reasons: long-tail facts may be lost because preserving them contributes little to the training objective, whereas some popular facts may be altered because their existing associations directly interfere with learning the target update. This hypothesis opens a path toward predicting factual damage before updating and prioritizing preservation accordingly.
  
\section*{Limitations}

Our popularity measure is based on the in-degree of the object entity and therefore captures entity-level structural popularity rather than the frequency of the complete factual proposition $(s,r,o)$. Although its positive correlations with Wikipedia frequency and pageviews provide external support for this proxy, relation frequency and other properties of the full triple may also contribute to the observed vulnerability. Future work should disentangle entity connectivity, relation rarity, and proposition-level frequency through relation-controlled analyses and broader corpus-based measurements.

Our main analysis uses controlled object substitutions and LoRA-based fine-tuning with a fixed optimization configuration. This setting supports comparisons across update targets, but may not capture the behavior of full-parameter fine-tuning, longer continual-training streams, or other knowledge-updating methods. Future work should test whether structural popularity remains predictive under more diverse update objectives, data scales, and optimization procedures.


\section{Ethical Considerations}
Knowledge updating can alter factual behavior beyond the intended target, potentially introducing or amplifying misinformation when deployed without adequate validation. Although our counterfactual updates are used only as controlled experimental interventions, similar techniques could be misused to manipulate model knowledge. Practical applications should therefore verify both update success and collateral effects, particularly in high-stakes domains, and retain human oversight over consequential updates.

Our factual graph is derived from Wikipedia and Wikidata and may inherit their coverage gaps, annotation errors, and societal biases. In addition, prioritizing structurally popular knowledge for preservation could further favor well-represented entities while providing less protection to long-tail knowledge. We view structural popularity as a diagnostic and mitigation signal rather than a universal measure of factual importance. Future systems should combine it with signals reflecting reliability, domain risk, and underrepresented knowledge to avoid reinforcing existing imbalances.
\bibliography{custom}
\appendix
\section{\textsc{FactProp} Construction and Validation}
\label{app:factprop}

\subsection{Grounded Graph Construction}

We construct the factual graph $\mathcal{G}_{\mathrm{fact}}$ using an automated expansion-and-filtering pipeline. Each node is an entity, and each directed edge is a verified factual triple $(s,r,o)$ from subject entity $s$ to object entity $o$ under relation $r$. Facts are connected when their subject or object entities overlap, and paths through these shared entities define the graph distances used in our evaluation.

Starting from high-confidence seed triples, such as \texttt{("Minecraft", "DevelopedByPrimary", "Mojang Studios")}, we iteratively expand from the entities collected so far using a predefined set of factual relations. During expansion, we discard a candidate triple if its object entity has already appeared earlier on the same traversal path. This prevents paths from revisiting ancestor entities and removes trivial cycles. Candidate triples are subsequently filtered to remove unsupported, ambiguous, or non-primary objects.

We use the DeepSeek API (\texttt{deepseek-chat}; accessed August 2026)~\cite{deepseekapi2026} to propose candidate triples and generate their corresponding natural-language questions. The full generation process consumed approximately $309.2$ million tokens, including $182.6$ million prompt tokens and $126.6$ million completion tokens. Each candidate triple is then verified against Wikidata~\cite{vrandecic2014wikidata} through an external validation module, and unverified candidates are discarded. The resulting graph contains $100{,}015$ unique entity nodes and $432{,}562$ verified relation edges spanning $39$ relation types.

\subsection{Relation and Answer Constraints}

Each QA item is grounded in a factual triple $(s,r,o)$. To determine whether a model prediction changes from correct to incorrect after updating, the expected answer to each query must be unambiguous. This condition is difficult to satisfy for one-to-many relations: for a subject--relation pair with multiple valid objects, a prediction may differ from the annotated object while remaining factually correct.

We therefore retain only subject--relation pairs $(s,r)$ with a unique or clearly primary expected object $o$. Relations such as \textit{CapitalOf} and \textit{DevelopedByPrimary} satisfy this requirement, whereas relations such as \textit{HasChild} are excluded because they may admit multiple equally valid answers. This constraint ensures that an observed mismatch reflects factual corruption rather than incomplete annotation. This design differs from resources such as MQUAKE~\cite{zhong2023mquake} and RippleEdits~\cite{cohen2023ripple}, which evaluate multi-hop or downstream consequences of model edits without explicitly enforcing answer cardinality for every queried subject--relation pair.

\subsection{QA Generation and Evaluation Control}

Each retained factual triple is verbalized into a natural-language question whose expected answer is the object entity $o$. The generated question is stored directly as an attribute of the corresponding graph edge. Model training and evaluation use these natural-language questions rather than serialized triples.

For each factual item, the exact same question string is used before and after the knowledge update. This controls for prompt variation and ensures that observed prediction changes are not caused by differences in wording. Defining factual connections through grounded triples also avoids relying on question-level surface similarity, since the same fact can be expressed in different ways and similar question templates can correspond to unrelated facts.

\subsection{External Validation of the Popularity Proxy}

For a factual triple $(s,r,o)$, we use the in-degree of object entity $o$ as its graph-based popularity score. In-degree counts how many verified factual relations point to the same entity and therefore measures how broadly the answer entity is referenced across the factual graph.

Because this measure is defined within $\mathcal{G}_{\mathrm{fact}}$, we compare it with two external indicators of entity popularity on the QID-resolved subset of the graph: (i) entity surface-form frequency in $200{,}000$ English Wikipedia articles, comprising approximately $100$ million tokens and matched using an Aho--Corasick automaton; and (ii) total 2024 Wikipedia pageviews for the corresponding entity article, restricted to user-agent traffic.

After retaining entities for which all three signals are non-zero, the evaluation set contains $35{,}868$ entities, corresponding to $59.9\%$ of the $59{,}932$ QID-resolved nodes. Table~\ref{tab:popularity_proxy_validation} reports both log--log Pearson correlations and rank-based Spearman correlations. In-degree is positively associated with Wikipedia frequency (Pearson $r=0.413$; Spearman $\rho=0.308$) and pageviews (Pearson $r=0.260$; Spearman $\rho=0.235$). Pageviews and Wikipedia frequency are also positively associated (Pearson $r=0.335$; Spearman $\rho=0.351$). All correlations are statistically significant.

\begin{table}[t]
\centering
\small
\setlength{\tabcolsep}{4pt}
\begin{tabular}{lrr}
\toprule
\textbf{Signal Pair} & \textbf{Pearson $r$} & \textbf{Spearman $\rho$} \\
\midrule
In-degree / Wiki frequency & 0.413 & 0.308 \\
In-degree / Pageviews & 0.260 & 0.235 \\
Pageviews / Wiki frequency & 0.335 & 0.351 \\
\bottomrule
\end{tabular}
\caption{\textbf{External validation of graph in-degree as a popularity proxy.} Pearson correlations are computed in log--log space; Spearman correlations compare ranks. All pairs use the same $35{,}868$ entities with non-zero values for all three signals.}
\label{tab:popularity_proxy_validation}
\end{table}

\begin{figure*}[t]
    \centering
    \includegraphics[width=\textwidth]{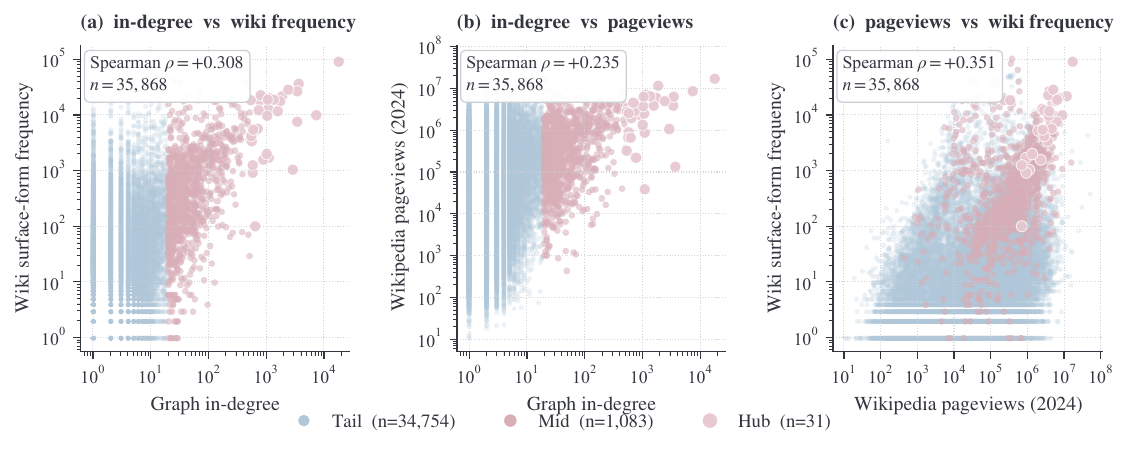}
    \caption{\textbf{Relationship between graph connectivity and external popularity signals.} The panels compare object in-degree, entity surface-form frequency in English Wikipedia, and 2024 Wikipedia pageviews. The consistently positive but moderate correlations support in-degree as a related, non-redundant proxy for factual popularity.}
    \label{fig:popularity_proxy}
\end{figure*}

The agreement is strongest at the head of the distribution: among the top-$10$ entities ranked by Wikipedia frequency, $8$ also fall in the top decile by in-degree, and \textit{United States} (Q30) is the maximum on both axes. The disagreement is concentrated in complementary regimes. Entities such as \textit{ESPN}, \textit{Dell Technologies}, and \textit{Assembly language} are densely interlinked in $\mathcal{G}_{\mathrm{fact}}$ (in-degree $\geq 15$) but occur only once in the sampled Wikipedia text, whereas some entities with high pageviews have relatively sparse factual annotations in the graph. These differences are informative rather than evidence that the signals are interchangeable: in-degree measures how many distinct verified factual relations resolve to an entity, while textual frequency and pageviews reflect corpus occurrence and public attention. We therefore treat the latter two signals as external corroboration rather than substitutes for graph in-degree.

\section{Additional Experimental Details}
\label{app:experimental_details}

\subsection{Update Data Construction}

For each target update $(s,r,o')$, we first generate $30$ QA templates that express the same subject--relation pair using different question forms. We then apply paraphrase augmentation to obtain $150$ targeted update examples. The final update set contains $650$ QA pairs: these $150$ targeted examples, $400$ neutral factual QA pairs sampled from graph-distant regions, and $100$ out-of-domain QA pairs. The neutral and out-of-domain examples reduce overfitting to the target association and help preserve unrelated model behavior. The out-of-domain examples are disjoint from the unrelated-knowledge evaluation set.

\subsection{LoRA Optimization}

We implement each factual update using LoRA~\cite{hu2021lora}. LoRA adapters are applied to the \texttt{q\_proj} and \texttt{v\_proj} matrices, while the base-model parameters remain frozen. Unless otherwise specified, we use LoRA rank $r=20$, scaling factor $\alpha=40$, and dropout $p=0.1$. We optimize the adapters with AdamW~\cite{loshchilov2019decoupled}, using a learning rate of $2.5\times10^{-4}$, batch size $4$, and up to $8$ epochs. Training stops early when the loss falls below $10^{-3}$.

\subsection{Injection Success and Answer Evaluation}

For each target query $q_{s,r}$, we construct a candidate set $\mathcal{C}_{s,r}$ containing the injected object $o'$ and comparison objects. For a candidate sequence $y=(y_1,\ldots,y_T)$, we use the joint log-probability
\begin{equation}
S_\theta(y\mid q_{s,r})=
\sum_{t=1}^{T}\log P_\theta(y_t\mid q_{s,r},y_{<t}).
\end{equation}
An injection is considered successful when $o'$ receives the highest score in $\mathcal{C}_{s,r}$. This candidate-scoring protocol evaluates whether the target association was learned without conflating injection success with open-ended decoding behavior. We retain both all-attempted and successful-update-only views when diagnosing the effects of failed updates, but make no aggregate successful-only claim unless the corresponding result is explicitly reported.

For non-target factual evaluation, we use unconstrained greedy decoding and alias-normalized, case-insensitive exact match after stripping punctuation. The same QA prompt is used before and after updating, and accepted aliases of the same Wikidata entity are mapped to a common answer.

\subsection{Subtree-Constrained Neighbor Sampling}

Candidate neighborhoods are expanded under a subtree constraint. At hop $d$, expansion begins only from entities selected at hop $d-1$; consequently, an evaluated path remains continuous from the update source instead of combining independently sampled nodes from unrelated branches. During final evaluation, we sample up to $30$ non-target facts without replacement at each distance $d\in\{1,\ldots,5\}$. Only facts answered correctly by the base model are retained for Flip Rate computation.

\subsection{Unrelated-Knowledge Control}

The unrelated evaluation set contains $50$ factual questions that do not occur in the targeted, neutral, or out-of-domain update data. These questions are also selected outside the evaluated target neighborhoods. We use this set only as a control for broad post-update degradation and do not include it in the Flip Rate calculation.

\subsection{Public-Benchmark Update Details}

For CounterFact and MQuAKE-CF, we construct one entity-disjoint batch of $100$ updates per benchmark. All strategies use the same LoRA objective and the same $N=100$ anchor budget. Evaluation facts are answered correctly by the base model and are entity-disjoint from both the updates and anchor sets. Popular and Rare Anchoring use the highest- and lowest-in-degree strata, respectively; Similarity Anchoring ranks candidate questions by cosine similarity to update questions using \texttt{all-MiniLM-L6-v2}. Random Anchoring samples uniformly from the eligible candidate pool. In \textsc{FactProp}, this pool excludes hub facts so that the sampled anchors track the mean popularity of the non-hub pool.

\section{Surface-Similarity Diagnostics}
\label{app:semantic_diagnostics}

\subsection{Four-Model Pooled Analysis}

To support the broad similarity analysis in Section~\ref{subsec:semantic_control}, we report the Pearson correlation between normalized Levenshtein similarity (source versus neighbor entity name) and binary flip status for each model on the pre-update-correct pool. As shown in Table~\ref{tab:permodel_pooled_r}, $r$ ranges from $0.026$ to $0.072$, with $r^2<0.6\%$ in every model. Although the large sample size makes each correlation statistically significant, the effect sizes are practically negligible.

\begin{table*}[t]
\centering
\small
\begin{tabular}{lrrrrr}
\toprule
\textbf{Model} & \textbf{$n$} & \textbf{Flip Rate (\%)} & \textbf{Pearson $r$} & \textbf{$r^2$ (\%)} & \textbf{$p$-value} \\
\midrule
Qwen3.5-2B      & 28{,}267 & 37.72 & 0.0340 & 0.12 & $1.05 \times 10^{-8}$ \\
Qwen3.5-9B      & 38{,}177 & 52.36 & 0.0719 & 0.52 & $6.74 \times 10^{-45}$ \\
Qwen3.6-27B     & 31{,}948 & 33.62 & 0.0259 & 0.07 & $3.65 \times 10^{-6}$ \\
Gemma-4-31B-it  & 40{,}587 & 21.31 & 0.0690 & 0.48 & $5.71 \times 10^{-44}$ \\
\midrule
\textbf{Pooled} & \textbf{138{,}979} & \textbf{36.01} & \textbf{0.0502} & \textbf{0.25} & \textbf{$3.39 \times 10^{-78}$} \\
\bottomrule
\end{tabular}
\caption{\textbf{Per-model correlation between surface similarity and flip status.} The statistically detectable correlations explain less than $0.6\%$ of outcome variance in every model.}
\label{tab:permodel_pooled_r}
\end{table*}

\subsection{Within-Neighborhood Paired Control}

Aggregate results may be influenced by differences across update sources and graph distances. We therefore group neighbors by both update source and hop distance and compute the correlation between similarity and flip status within each $(\text{source},d)$ group. Groups with fewer than three samples or constant similarity or flip status are excluded. Table~\ref{tab:paired_r_per_model} shows that the mean and median within-group correlations are near zero and that positive and negative signs are approximately balanced.

\begin{table*}[t]
\centering
\small
\begin{tabular}{lrrrrr}
\toprule
\textbf{Model} & \textbf{$n_{\mathrm{groups}}$} & \textbf{Mean $r$} & \textbf{Median $r$} & \textbf{$r>0$} & \textbf{$r<0$} \\
\midrule
Qwen3.5-2B      & 158 & $+0.0191$ & $+0.0191$ & 88 & 70 \\
Qwen3.5-9B      & 159 & $+0.0037$ & $+0.0036$ & 83 & 76 \\
Qwen3.6-27B     & 117 & $-0.0335$ & $-0.0149$ & 54 & 63 \\
Gemma-4-31B-it  & 154 & $+0.0027$ & $+0.0105$ & 88 & 66 \\
\midrule
\textbf{Pooled} & \textbf{588} & \textbf{$+0.0002$} & \textbf{$+0.0061$} & \textbf{313} & \textbf{275} \\
\bottomrule
\end{tabular}
\caption{\textbf{Paired Pearson correlations within fixed source--hop neighborhoods.} Surface similarity carries no systematic signal once the update source and graph distance are fixed.}
\label{tab:paired_r_per_model}
\end{table*}

Across the $588$ valid groups, source--neighbor pairs with Levenshtein similarity of at least $0.5$ account for only approximately $3\%$ of the controlled evaluation pool. This confirms that highly similar pairs are too sparse to explain the broad error pattern.

\subsection{Secondary Paired Audit}

For completeness, we retain the earlier, narrower paired audit over five individual source reports in Table~\ref{tab:semantic_corr_secondary}. This diagnostic predates the four-model pooled analysis and is treated as secondary evidence rather than as the basis of the main claim. Its correlations are weak and mixed in sign, ranging from $-0.09$ to $+0.06$, with only eight high-similarity examples in the clean-correct subset.

\begin{table*}[t]
\centering
\small
\begin{tabular}{llrrrr}
\toprule
\textbf{Report} & \textbf{Relation} & \textbf{Raw $r$} & \textbf{Clean-correct $r$} & \textbf{Raw high-$n$} & \textbf{Clean high-$n$} \\
\midrule
Hub\_Sample\_1 & CountryOfCity & -0.0488 & -0.0908 & -- & -- \\
Low\_Sample\_1 & CountryOfCity & -0.0742 & -0.0812 & -- & -- \\
Hub\_Sample\_2 & CountryOfInc. & -0.0069 & 0.0525 & -- & -- \\
Low\_Sample\_2 & CountryOfInc. & 0.0612 & -0.0012 & -- & -- \\
Low\_Sample\_3 & CountryOfInc. & -0.0078 & 0.0190 & -- & -- \\
\midrule
\textbf{Mean} & -- & \textbf{-0.0153} & \textbf{-0.0203} & \textbf{35 total} & \textbf{8 total} \\
\bottomrule
\end{tabular}
\caption{\textbf{Secondary per-report similarity audit.} The earlier narrow audit also finds negligible correlation between lexical proximity and flip likelihood.}
\label{tab:semantic_corr_secondary}
\end{table*}

\section{Additional Popularity-Anchoring Results}
\label{app:mitigation_by_source}

On \textsc{FactProp}, Popularity Anchoring selects prompts whose answer entities fall in the high object-in-degree stratum. Random Anchoring samples uniformly from the non-hub pool, and Rare Anchoring selects from its lowest-in-degree stratum. All anchored strategies use $N=100$ prompts and the same KL behavior-preservation objective; No Anchoring omits that regularizer. We separately report Popular-, Average-, and Rare-source updates to verify that the aggregate mitigation advantage is not driven by one type of target. Table~\ref{tab:mitigation_by_source} shows that Popular Anchoring achieves the lowest average Flip Rate in all three source groups.

\begin{table}[H]
\centering
\small
\resizebox{\columnwidth}{!}{%
\begin{tabular}{lcccccc}
\toprule
\textbf{Method} & \textbf{d1} & \textbf{d2} & \textbf{d3} & \textbf{d4} & \textbf{d5} & \textbf{Avg.} \\
\midrule
\multicolumn{7}{l}{\textit{Popular-source updates ($n{=}10$ targets)}} \\
No Anchoring & 90.9 & 85.9 & 77.0 & 74.9 & 73.6 & 80.5 \\
Random Anchoring & \textbf{83.7} & 80.6 & 73.5 & 70.3 & 73.4 & 76.3 \\
Rare Anchoring & 89.5 & 80.3 & 66.1 & 66.5 & \textbf{69.3} & 74.4 \\
Popular Anchoring & 84.1 & \textbf{75.7} & \textbf{65.8} & \textbf{64.0} & 69.5 & \textbf{71.8} \\
\midrule
\multicolumn{7}{l}{\textit{Average-source updates ($n{=}10$ targets)}} \\
No Anchoring & 92.9 & 67.7 & 79.0 & 72.8 & 73.0 & 77.1 \\
Random Anchoring & 85.7 & 63.9 & 75.3 & 69.6 & 70.4 & 73.0 \\
Rare Anchoring & 90.6 & 61.1 & 69.0 & 64.7 & 63.7 & 69.8 \\
Popular Anchoring & \textbf{82.1} & \textbf{60.2} & \textbf{68.3} & \textbf{59.5} & \textbf{61.6} & \textbf{66.4} \\
\midrule
\multicolumn{7}{l}{\textit{Rare-source updates ($n{=}10$ targets)}} \\
No Anchoring & 100.0 & 85.3 & 81.6 & 81.1 & 77.4 & 85.1 \\
Random Anchoring & \textbf{75.0} & 82.8 & 76.3 & 75.1 & 71.8 & 76.2 \\
Rare Anchoring & 100.0 & 75.2 & \textbf{68.1} & 68.0 & 62.6 & 74.8 \\
Popular Anchoring & 75.0 & \textbf{62.4} & 74.4 & \textbf{66.7} & \textbf{60.7} & \textbf{67.8} \\
\bottomrule
\end{tabular}%
}
\caption{\textbf{Mitigation Breakdown by Update Source Group (Flip Rate \% across source strata).} Popular Anchoring achieves the lowest average Flip Rate for Popular-, Average-, and Rare-source updates in this evaluation.}
\label{tab:mitigation_by_source}
\end{table}

\section{Attention Perturbation Diagnostics}
\label{app:attention_diagnostics}

This section provides hop-wise supporting evidence for the attention analysis in Section~\ref{subsec:mechanisms}. We report the three completed paired audits: Qwen3.5-2B, Qwen3.5-9B, and Gemma-4-31B-it.

\paragraph{Attention Measurement Details.}
For each neighboring query, we collect generation attentions at the first decoding step, keep the final full-attention layer, and average over attention heads and query positions. The evaluated span is obtained by tokenizing the queried entity string and locating that subsequence in the prompt. We sum the attention mass on this span and normalize it by the span-length baseline $|S|/K$. We report the absolute clean-to-updated change, $|\Delta\mathrm{AttLift}|$, on the clean-correct subset.

\begin{table*}[t]
\centering
\small
\setlength{\tabcolsep}{4pt}
\begin{tabular}{llrrrrr}
\toprule
\textbf{Model} & \textbf{Source} & \textbf{d1} & \textbf{d2} & \textbf{d3} & \textbf{d4} & \textbf{d5} \\
\midrule
\multirow{2}{*}{Qwen3.5-2B}
 & Popular & \textbf{0.956} & 0.476 & 0.468 & 0.534 & 0.524 \\
 & Rare    & 0.851 & \textbf{0.754} & \textbf{0.584} & \textbf{0.564} & \textbf{0.597} \\
\midrule
\multirow{2}{*}{Qwen3.5-9B}
 & Popular & \textbf{0.421} & \textbf{0.334} & \textbf{0.284} & \textbf{0.277} & 0.242 \\
 & Rare    & 0.187 & 0.311 & 0.274 & 0.255 & \textbf{0.273} \\
\midrule
\multirow{2}{*}{Gemma-4-31B-it}
 & Popular & \textbf{0.139} & 0.083 & \textbf{0.096} & \textbf{0.106} & \textbf{0.102} \\
 & Rare    & 0.102 & \textbf{0.093} & 0.076 & 0.071 & 0.076 \\
\bottomrule
\end{tabular}
\caption{\label{tab:attention_lift_masked} \textbf{Hop-wise attention perturbation in the three completed paired audits.} Each cell reports mean $|\Delta\mathrm{AttLift}|$ on the clean-correct subset; bold marks the larger source class within each model and hop. Popular-source updates produce the larger perturbation at $d{=}1$ in all three models, whereas later-hop comparisons are mixed.}
\end{table*}

Table~\ref{tab:attention_lift_masked} shows that Popular-source updates produce larger perturbations at $d=1$ in all three models, which is the consistent pattern summarized in the main text. Later-hop comparisons are mixed: Rare-source updates are larger at most distant hops for Qwen3.5-2B, while the ordering varies by hop for Qwen3.5-9B and Gemma-4-31B-it. We therefore treat attention perturbation as a diagnostic associated with the immediate-neighbor pattern, not as evidence of a universal causal mechanism across all distances.

\end{document}